\documentclass{nature}

\usepackage{times}
\usepackage{graphicx} 
\usepackage{subfigure} 

\usepackage{hyperref}
\usepackage{amsmath}
\usepackage{amsfonts}
\usepackage{amssymb}

\def\<{\begin{equation}}
\def\>{\end{equation}}

\usepackage[compress]{natbib}
\bibpunct{}{}{,}{s}{}{\textsuperscript{,}}
\renewcommand\bibnumfmt[1]{#1.}

\title{The Wilson Machine for Image Modeling}

\author{Saeed Saremi$^{1}$ \& Terrence J. Sejnowski$^1$}

\begin{document}

\maketitle

\begin{affiliations}
 \item Salk Institute, 10010 N Torrey Pines Rd, La Jolla, CA 92037.
\end{affiliations}

\begin{abstract}
Learning the distribution of natural images is one of the hardest and most important problems in machine learning, computer vision, and artificial intelligence~\citep{lecun2015deep, schmidhuber2015deep, gerhard2015modeling}. Intelligent agents must navigate in the natural world; having the right model of natural images is thus essential for their success in visual tasks \citep{ullman2010vision,simoncelli2001natural}. The problem remains open, however, because the enormous complexity of the structures in natural images spans all length scales~\citep{simoncelli2001natural, mandelbrot1983fractal, wilson1979problems}. Here we break down the complexity of the problem and show that the hierarchy of structures in natural images fuels a new class of learning algorithms based on the theory of critical phenomena and stochastic processes.  We approach this problem from the perspective of the theory of critical phenomena, which was developed in condensed matter physics to address problems with infinite length-scale fluctuations~\citep{wilson1979problems, wilson1974renormalization, ma2000modern}, and build a framework to integrate the criticality of natural images~\citep{saeed13-1, saeed14-1, saeed15-1, bialek13-1} into a learning algorithm. The problem is broken down by mapping images into a hierarchy of binary images, called bitplanes. In this representation, the top bitplane is critical, having fluctuations in structures over a vast range of scales. The bitplanes below go through a gradual stochastic heating process to disorder. We turn this representation into a directed probabilistic graphical model~\citep{koller2009probabilistic}, transforming the learning problem into the unsupervised learning of the distribution of the critical bitplane and the supervised learning of the conditional distributions for the remaining bitplanes. We learnt the conditional distributions by logistic regression in a convolutional architecture. Conditioned on the critical binary image, this simple architecture can generate large, natural-looking images, with many shades of gray, without the use of hidden units, unprecedented in the studies of natural images. The framework presented here is a major step in bringing criticality and stochastic processes to machine learning and in studying natural image statistics.
\end{abstract}

Learning the distribution of natural images remains one of the hardest problems in unsupervised learning. Structures in natural images are complex, varied, and most importantly span a vast range of length scales, from the smallest within a few dozen pixels to large structures the size of the image itself. In this respect, they are very similar to critical points realized for physical systems near continuous (second-order) phase transitions~\citep{wilson1979problems}. There is a deep connection underlying this similarity as critical large-scale fluctuations emerge after mapping natural images to a stack of binary images~\citep{saeed13-1, saeed14-1, saeed15-1}. Here we introduce  an algorithm to utilize this criticality in learning the distribution of natural images.

The binary representation of Ref. \citenum{saeed13-1} contained ordered and disordered phases, with a critical bitplane close to a phase transition. The critical bitplane underlies the large-scale fluctuations of the structures in natural images. Here we build on those results and exploit the binary representation, turning it into a directed probabilistic graphical model~\citep{koller2009probabilistic} by placing the critical bitplane at the root of the directed graph. This transforms learning the distribution of natural images into the \emph{unsupervised} learning of the distribution of the critical bitplane and the \emph{supervised} learning of the conditional distributions of all other bitplanes in a parent-child hierarchy outlined below. The network architecture for learning is simple. We approximate the conditional distributions by logistic regression, where the weights to the children nodes are learnt in a convolutional architecture by weight-sharing.

Turning unsupervised learning into supervised learning has a rich history and goes back to the wake-sleep algorithm and the Helmholtz machine, which co-trained inference (the wake phase) and generative models (the sleep phase) together~\citep{hinton1995wake, dayan1995helmholtz}. Here, in contrast, the supervision is given by the input itself. Instead of learning deep architectures with hidden units~\citep{lecun2015deep,schmidhuber2015deep}, we show that the visible hierarchy in the form of critical structures and stochastic processes can be utilized for learning. These two types of hierarchies complement each other. In particular, the visible hierarchy can be modeled using hidden units in the deep learning framework.

The starting point is to construct an \emph{exact} binary representation so that the top bitplane is critical and the bottom ones go through a stochastic heating process to disorder. In the binary representation that was studied in \citenum{saeed13-1, saeed15-1}, the critical bitplane was in the middle of the hierarchy, with different types of structures emerging in relation to it, ``ordered'' towards the top, and ``disordered'' towards the bottom. It is not clear in that representation how the bitplanes just above and just below the critical bitplane could influence each other in a directed graph. However, in the new representation, the causal direction is clear:  the stochastic heating process from criticality (top) to disorder (bottom).

The new representation is a variation on the original~\citep{saeed13-1} but it turns out to be key in how we map the binary representation to a directed graphical model. We first give a brief review of the original representation. The analog values $I$ (assumed to be non-negative integers for simplicity) were mapped to a binary representation $\{B_1, B_2, \cdots, B_\Lambda \}$ by the following decomposition \<  I = \sum_{\lambda=1}^{\Lambda} 2^{\Lambda-\lambda} \ B_\lambda,\>
where $B_{\lambda} \in \{0,1\}$ were found iteratively starting from $\lambda=1$. The visual representation of the binary map from analog values to their corresponding bits is given in Fig.~\ref{fig:analog2bin}. The binary map is extended to any of array, mapping the gray-scale image $\mathcal{I}$ to a stack of binary images $\mathbb{B} = \{\mathcal{B}_1, \mathcal{B}_2, \cdots, \mathcal{B}_\Lambda\}.$

The goal is to build a directed probabilistic graphical model on top of this representation, where 
\<\label{eq:directed} P(\mathbb{B}) = \newline P(\mathcal{B}_1) P(\mathcal{B}_2| \mathcal{B}_1) \cdots P(\mathcal{B}_\Lambda|\mathcal{B}_1, \mathcal{B}_2, \cdots, \mathcal{B}_{\Lambda-1}).\> However, to find good estimates for the conditional distributions, the structure of the directed graph should be dictated by a ``physical'' causal process. In a simple tweak, we change the binary representation so that $\mathcal{B}_1$ is critical and the bitplanes below go through a gradual stochastic process to disorder. The stochastic process becomes the causal direction, and the parent-child hierarchy becomes clear: from criticality to disorder.

In the new representation, all the half-point dividing lines in Fig.~\ref{fig:analog2bin} are changed to the median values of their corresponding intervals. This hierarchical median-thresholding is \emph{exactly} equivalent to applying \emph{half-point} thresholding of Fig.~\ref{fig:analog2bin} after first performing histogram equalization \< I\rightarrow I' = \int_{-\infty}^{I} P(x) dx,\> where $P(x)$ is the marginal distribution of the pixel intensities. The half-point thresholding of the histogram-equalized image is equivalent to median-thresholding of the original image because histogram equalization does not change the rank of pixel values. In fact, any strictly monotonic transformation of $\mathcal{I}$ yields the same binary representation $\mathbb{B}$ under median-thresholding. The illustration of this new binary representation for an image in the Geisler database~\citep{geisler2011statistics} is given in Fig.~\ref{fig:bitplanes}.

The first bit in the new representation $\mathcal{B}_1 =  \mathcal{I}>{\tt median}(\mathcal{I}) $ is the median-thresholded image, which was studied at length in~\citenum{saeed14-1, saeed15-1, bialek13-1}. In particular, it was shown that $\mathcal{B}_1$ is at the onset of a percolation transition, with connected clusters governed by scaling laws on all length scales ~\citep{saeed15-1}. In fact, these rich connected clusters at the top of the graphical model provide the fuel to the algorithm presented here. 

What we gain in the new representation is what we planned to achieve in that, as $\lambda$ increases, the bitplanes $\mathcal{B}_\lambda$ go through a gradual stochastic heating process to disorder  (see Fig.~\ref{fig:bitplanes}), thus placing them in the parent-child hierarchy of the directed graphical model of Eq.~\ref{eq:directed}. In summary, in the new representation, $\mathcal{B}_1 =  \mathcal{I}>{\tt median}(\mathcal{I})$ is critical, at the root of the graph, and $\mathcal{B}_\lambda$ is the parent of $\mathcal{B}_{\lambda'}$ for $\lambda<\lambda'$. The stochastic process from criticality to disorder \emph{is} the hierarchy, seen in Fig.~\ref{fig:bitplanes} by the islands of connected clusters gradually dissolving in a sea of noise. With this construction of the graphical model, rooted in statistical physics and in the theory of critical phenomena, we find  estimates for the conditional distributions $P(\mathcal{B}_\lambda|\mathcal{B}_1, \mathcal{B}_2, \cdots, \mathcal{B}_{\lambda-1})$.

The network architecture for learning the conditional distributions is very simple. We first assumed that pixels in the bitplane $\mathcal{B}_\lambda$ are independent conditioned on the bitplanes above. This might appear as a strong assumption for bitplanes close to $\mathcal{B}_1$ but since the critical fluctuations happen at \emph{infinite} scales, large structures in $\mathcal{B}_1$ induce marginal interactions in the bitplanes below. We estimated the conditional distributions $P(\mathcal{B}_\lambda|\mathcal{B}_1, \mathcal{B}_2, \cdots, \mathcal{B}_{\lambda-1})$ by taking $L\times L$ patches and performing logistic regression for bitplane $\mathcal{B}_\lambda$  taking bitplanes $\{\mathcal{B}_1, \mathcal{B}_2, \cdots, \mathcal{B}_{\lambda-1}\}$ as  input. Since natural images are translation-invariant, we assumed weight-sharing, making the logistic regression convolutional.


We performed the learning of the convolutional weights by maximizing the log likelihood with a single batch using the second-order Newton's method. The batch contained $10^5$ samples of size $41\times41$, taken from 1024 images of size $2844 \times 4284$ pixels from the Geisler database of natural images~\citep{geisler2011statistics}. The learnt center-surround receptive field for $P(\mathcal{B}_2|\mathcal{B}_1)$ is shown in Fig.~\ref{fig:rf}(a). Receptive fields for other conditional distributions have the same center-surround form. The results reported here were obtained by having $\ell_2$ prior on the weights; results were qualitatively similar, however, after setting the regularization term to $0$.

After learning the convolutional weights, conditioning on the first bitplane, we found the logistic activations of the lower bitplanes, up to $\mathcal{B}_8$. A key characteristic of natural images is the hierarchy in object sizes. This appears as connected clusters in the bitplane representation \citep{saeed15-1}. The peaks in Fig.~\ref{fig:logispeaks}(b) is the signature of those large clusters. To preserve the bigger clusters in lower bitplanes, which are key in natural images, we adopted a winner-take-all strategy, except for the interval $[0.4,0.6]$. In that interval we took samples according to the logistic activation. With this algorithm, we can easily generate binary images by the increasing order in $\lambda$, dictated by the structure of the directed graph. For example, the image in Fig.~\ref{fig:wow} was generated by conditioning on the bitplane $\mathcal{B}_1$ of Fig.~\ref{fig:bitplanes}, generating bitplanes $\mathcal{B}_2$ up to $\mathcal{B}_8$ with logistic activations, and combining $\{\mathcal{B}_1,\mathcal{B}_2,\cdots,\mathcal{B}_8\}$ to obtain the gray-scale image.

The model performance was evaluated by the normalized mean square error and the conditional log likelihood scores. The normalized mean square error for the generated images, averaged over the database is 0.0588. The same error measured for the null model, where the convolutional weights/biases are set to zero, is 0.1244.  For the density estimation measure, the average negative log likelihood for the conditional probabilities were evaluated and are given in Table~\ref{table:loglikehood}. Note that in this framework, the log likelihood for all bitplanes $\mathbb{B}$ combined is \emph{not} a good measure to compare models, in part because the analog image is obtained by a \emph{weighted} sum of the bitplanes. In other words, it is (much) more important to have a high log likelihood score for bitplanes closer to the critical bitplane $\mathcal{B}_1$ than the ones further below.

The state of the art in natural image modeling has been limited to texture synthesis, dead leaf images, small patches, and large images that do not look natural or realistic --- see Ref.~\citenum{gerhard2015modeling} for the most recent review. That being said, the direct comparison cannot be made to other models at the present stage, since the results here were obtained by conditioning on the critical bitplane $\mathcal{B}_1$ with its rich structures. Our results, however, point to a new research direction in image modeling, as learning the prior on the critical bitplane $\mathcal{B}_1$ will transform this algorithm to a fully probabilistic model and a very powerful model for natural images.


This novel framework is named after \emph{Kenneth G Wilson (1936 -- 2013)} for his immense contributions to our understanding of the nature of criticality. Here, the criticality of natural images plays the central role in the learning algorithm, and in contrast with the research focus in the deep learning community on learning deep architectures with hidden units, we showed that the hierarchy of \emph{visible} structures by itself fuels learning in the Wilson machine. 

Along these lines, there has been a recent work in learning data distributions by turning them into noise through stochastic processes, with the use of hidden units~\citep{sohl2015deep}. Learning in that framework is to figure out how to reverse the stochastic process, thus going from noise to data. In contrast, in the Wilson machine described here, the learning happens due to the stochastic process that is already present in the data, and is part of the data itself. In addition, the results here point to a new direction in image compression as learning a better probabilistic model in this framework might push the (lossy) compression limit to $1$ bit/pixel (i.e. the first bitplane $\mathcal{B}_1$.) Finally, even though we focused on natural images, the framework here is general and should be especially powerful for analog signals with very long correlation lengths/times.

\bibliography{wilsonmachine}
\bibliographystyle{nature}

\vspace{1cm}

\begin{addendum}
 \item We acknowledge conversations with Ruslan Salakhutdinov, comments from Lyle Muller on the manuscript, and the support of The Howard Hughes Medical Institute.

 \item[Correspondence] Correspondence and requests for materials
should be addressed to S.S.~(email: saeed@salk.edu).

\clearpage

\begin{figure}[h!]
\begin{center}
\includegraphics[width= 5.5 cm]{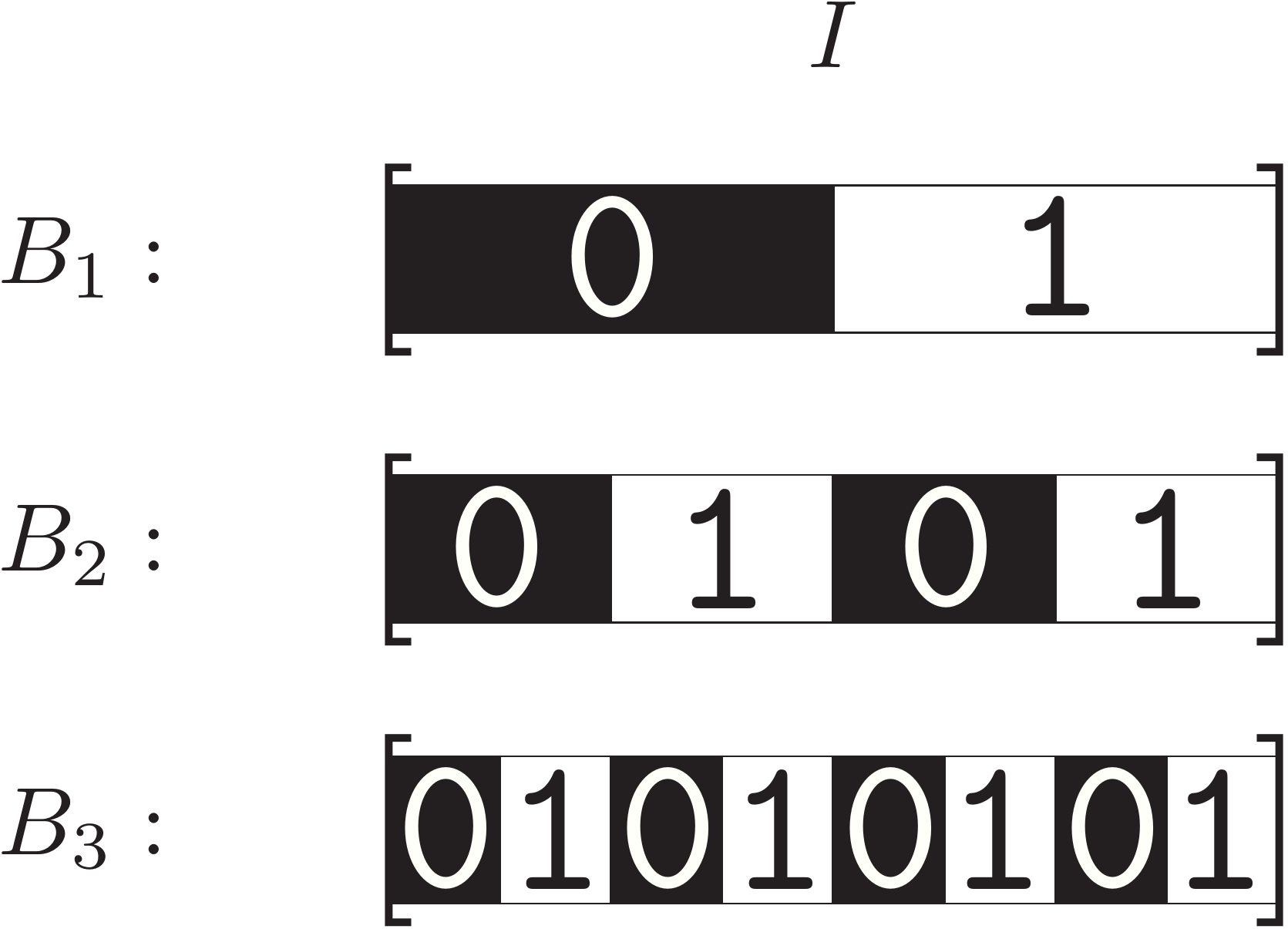}
\caption{\label{fig:analog2bin}  {\bf The binary representation of analog values.} The first three bits in the binary representation of the $\Lambda-$bit integer value $I$ in the range $[0,2^\Lambda-1]$ is shown schematically.  Each bit divides the interval in half iteratively, starting from the most significant bit $B_1$.}
\end{center}
\end{figure}

\begin{figure}[h!]
\begin{center}
\includegraphics[width= 0.65\textwidth]{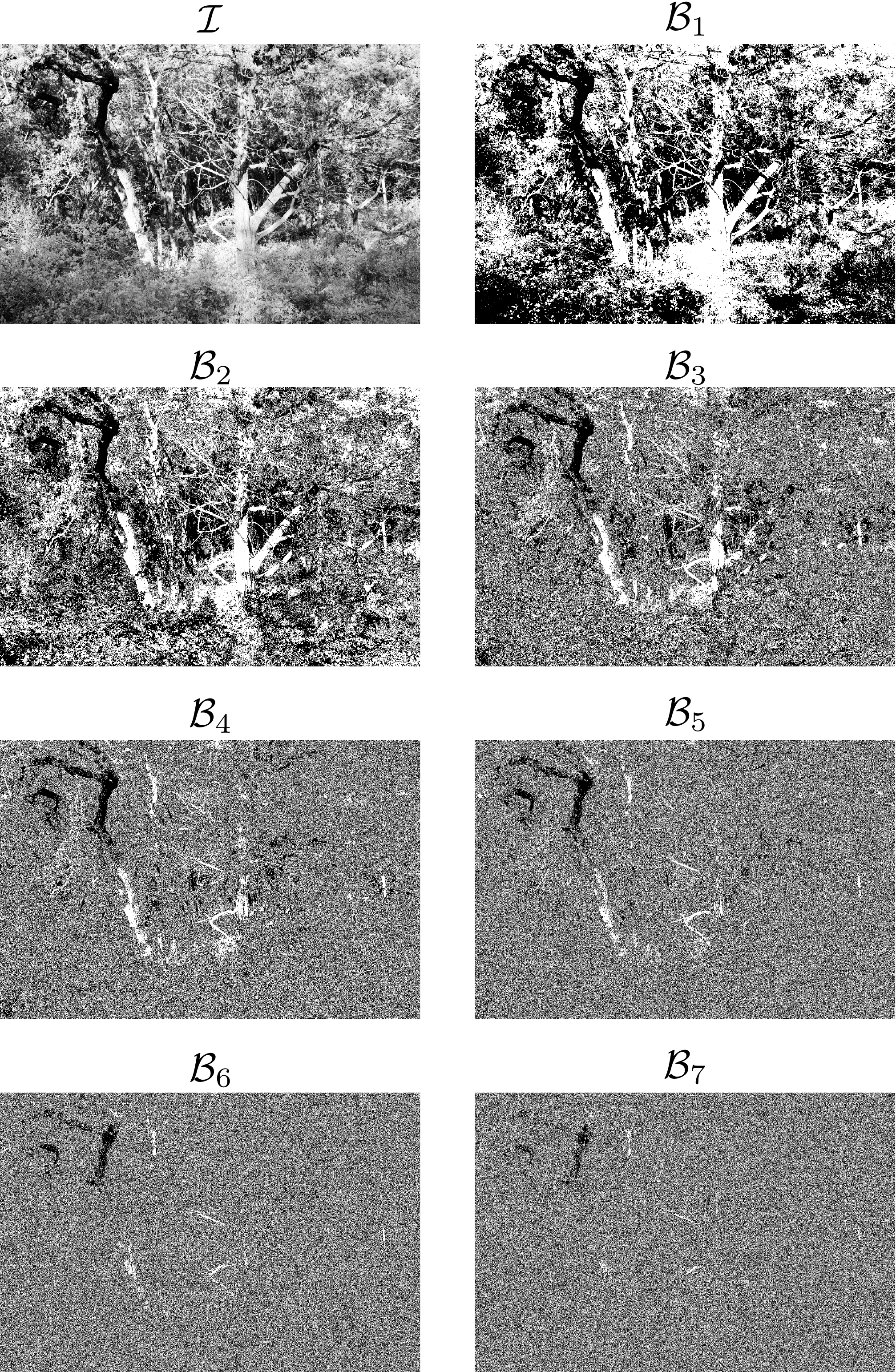}
\caption{\label{fig:bitplanes} {\bf Hierarchical median-thresholding, the critical bitplane, and the stochastic heating process to disorder.} An image $\mathcal{I}$ ($2844 \times 4284$ pixels) in the Geisler database of natural images and its binary representation up to bitplane $\mathcal{B}_7$ are shown. The critical bitplane $\mathcal{B}_1$ goes through a gradual stochastic heating process to disorder.}
\end{center}
\end{figure}

\begin{figure}[h!]
\begin{center}
\includegraphics[width= 0.65\textwidth]{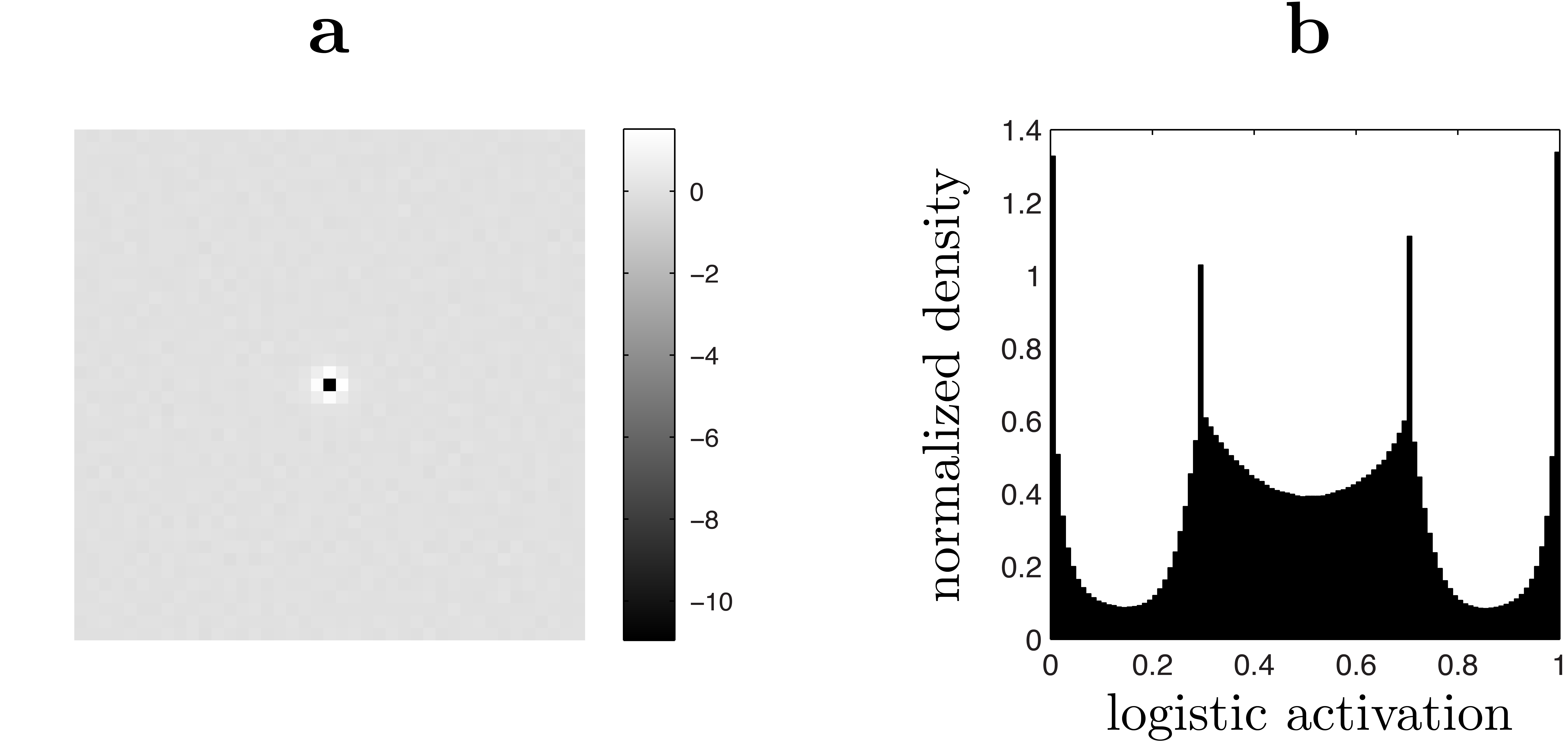}
\caption{\label{fig:rf} {\bf Learnt receptive fields and logistic activations.} {\bf a,} The plot of the learnt receptive field for the conditional distribution $P(\mathcal{B}_2|\mathcal{B}_1)$ is given. The size of the receptive field is $41\times41$ pixels. The receptive fields to other bitplanes have the same center-surround form. {\bf b,} \label{fig:logispeaks} The normalized histogram of the logistic activation function after applying the convolution filter  in (a) to the bitplane $\mathcal{B}_1$ from Fig.~\ref{fig:bitplanes} is plotted here. The peaks come from the large connected clusters in $\mathcal{B}_1$.}
\end{center}
\end{figure}

\begin{figure}[h!]
\begin{center}
\includegraphics[width= 0.65\textwidth]{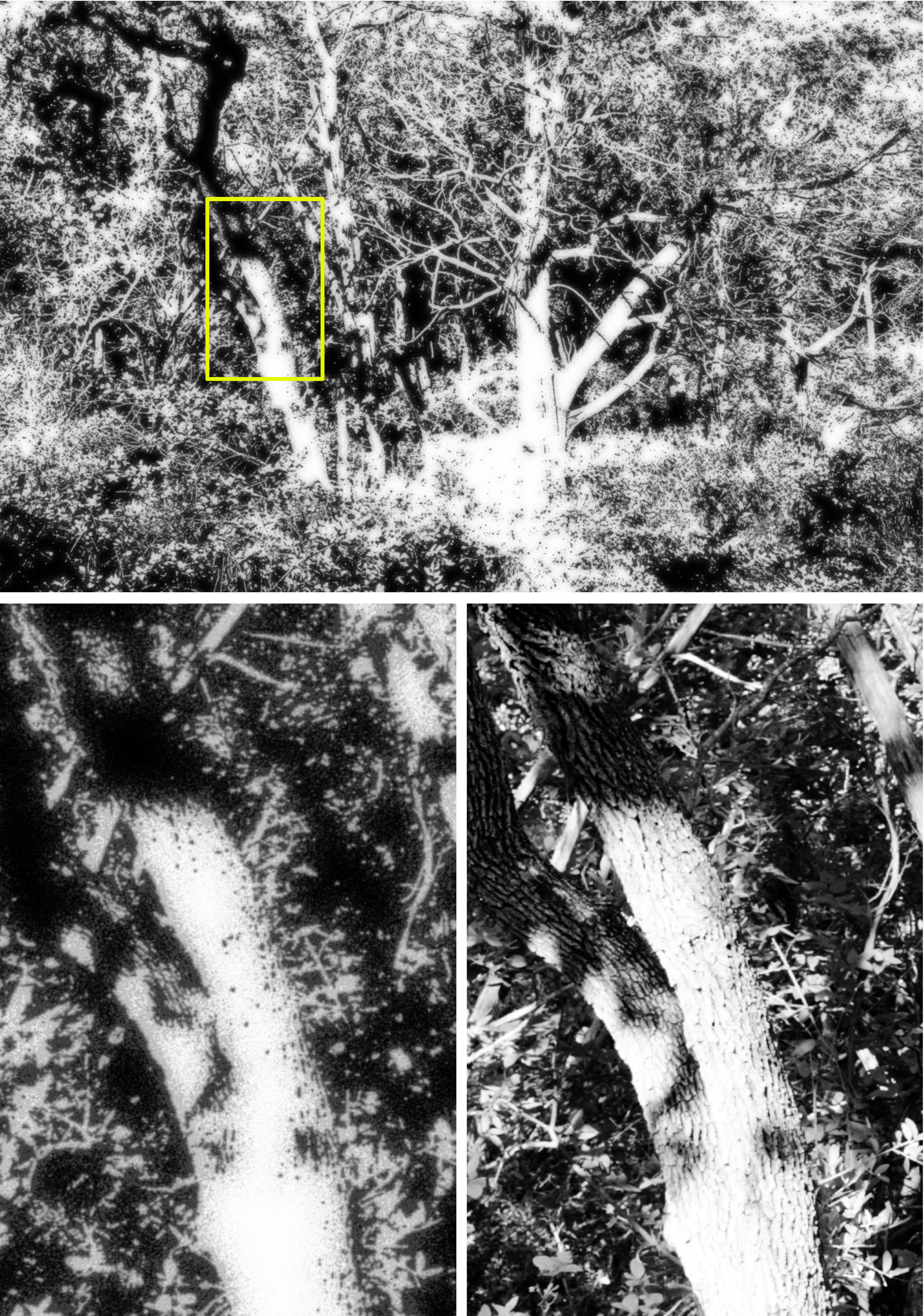}
\caption{\label{fig:wow} {\bf Image generation in the Wilson machine.} The image on top ($2564 \times 4004$ pixels) was generated by conditioning on the bitplane $\mathcal{B}_1$ of Fig.~\ref{fig:bitplanes} and obtaining other bitplanes up to $\mathcal{B}_8$ by the logistic activations from the learnt convolutional filters. The bitplanes were combined to obtain the gray-scale image. The area enclosed in the yellow rectangle is blown up on the left and the corresponding region in the original image is given on the right.}
\end{center}
\end{figure}

\clearpage

\begin{table}[t!]
\centering
\small
\begin{tabular}{|c|c|c|}\hline
${\rm NLL}(\mathcal{B}_2|\mathcal{B}_1)$ &${\rm NLL}(\mathcal{B}_3|\mathcal{B}_{1},\mathcal{B}_{2})$ &${\rm NLL}(\mathcal{B}_4|\mathcal{B}_{1},\mathcal{B}_{2},\mathcal{B}_{3})$  \\
\hline
$0.2369$ & $0.2901$ &$0.2749$\\
\hline
\end{tabular}
\caption{{\bf Density estimation results.} The results on the density estimation of the conditional distributions are given. The average negative log likelihood (NLL) is reported in bits/pixel. There were about $10^{10}$ pixels in the experiments. The negative log likelihood of the null model, corresponding to white noise, is 1 bit/pixel.}
\label{table:loglikehood}
\end{table}

\clearpage



\end{addendum}

\end{document}